\documentclass[conference]{IEEEtran}
\IEEEoverridecommandlockouts
\usepackage{cite}
\usepackage{amsmath,amssymb,amsfonts}
\usepackage{mathtools}

\usepackage{booktabs}

\usepackage{algorithm,algpseudocode}
\DeclareMathOperator*{\argmax}{argmax}

\usepackage{graphicx}
\usepackage{textcomp}
\usepackage{xcolor}
\usepackage[a4paper, total={184mm,239mm}]{geometry}

\newcommand{\ie}{\textit{i}.\textit{e}.}
\newcommand{\eg}{\textit{e}.\textit{g}.}

\makeatletter
\newcommand{\printfnsymbol}[1]{%
  \textsuperscript{\@fnsymbol{#1}}%
}
\makeatother

\makeatletter
\newcommand{\linebreakand}{%
  \end{@IEEEauthorhalign}
  \hfill\mbox{}\par
  \mbox{}\hfill\begin{@IEEEauthorhalign}
}
\makeatother

\usepackage[bookmarksnumbered=true]{hyperref} 
\hypersetup{
     colorlinks = true,
     linkcolor = black,
     anchorcolor = black,
     citecolor = black,
     filecolor = black,
     urlcolor = black
     }

\def\BibTeX{{\rm B\kern-.05em{\sc i\kern-.025em b}\kern-.08em
    T\kern-.1667em\lower.7ex\hbox{E}\kern-.125emX}}
\begin{document}

\title{Gradient-based Bit Encoding Optimization
for Noise-Robust Binary Memristive Crossbar
}

\author{
\IEEEauthorblockN{Youngeun Kim$^{*,1}$\thanks{\hspace{-3mm}$^*$ These authors have contributed equally to this work.}, Hyunsoo Kim$^{*,2}$, Seijoon Kim$^{2}$, Sang Joon Kim$^{2}$, Priyadarshini Panda$^{1}$} 
\thanks{\hspace{-3mm}$\dagger$ This work was carried out while Youngeun Kim worked as an intern at Samsung Advanced Institute of Technology, South Korea.} 
\IEEEauthorblockA{
$^{1}$Department of Electrical Engineering, Yale University, USA\\
$^{2}$Samsung Advanced Institute of Technology, South Korea\\
\{youngeun.kim, priya.panda\}@yale.edu, \{hs0128.kim, seijoon.kim, sangjoon0919.kim\}@samsung.com}
% \and
% \IEEEauthorblockN{Hyunsoo Kim$^{*}$} 
% \IEEEauthorblockA{
% % \textit{Department of Electrical Engineering} \\
% \textit{Samsung Advanced Institute of Technology}\\
% South Korea \\
% hs0128.kim@samsung.com }
% \and
% \IEEEauthorblockN{Seijoon Kim} 
% \IEEEauthorblockA{
% % \textit{Department of Electrical Engineering} \\
% \textit{Samsung Advanced Institute of Technology}\\
% South Korea \\
% seijoon.kim@samsung.com }
% \linebreakand
% \IEEEauthorblockN{Sang Joon Kim} 
% \IEEEauthorblockA{
% % \textit{Department of Electrical Engineering} \\
% \textit{Samsung Advanced Institute of Technology}\\
% South Korea \\
% sangjoon0919.kim@samsung.com}
% \and
% \IEEEauthorblockN{Priyadarshini Panda} 
% \IEEEauthorblockA{
% % \textit{Department of Electrical Engineering} \\
% \textit{Yale University}\\
% USA \\
% priya.panda@yale.edu}
}

\maketitle

\begin{abstract}
Binary memristive crossbars have gained huge attention as an energy-efficient deep learning hardware accelerator.
Nonetheless, they suffer from various noises due to the analog nature of the crossbars. 
To overcome such limitations, most previous works train  weight parameters with noise data obtained from a crossbar.
These methods are, however, ineffective because it is difficult to collect noise data in large-volume manufacturing environment where each crossbar has a large device/circuit level variation.
Moreover, we argue that there is still room for improvement even though these methods somewhat improve accuracy.
This paper explores a new perspective on mitigating crossbar noise in a more generalized way by manipulating input binary bit encoding rather than training the weight of networks with respect to noise data.   
We first mathematically show that the noise decreases as the number of binary bit encoding pulses increases when representing the same amount of information. 
In addition, we propose Gradient-based Bit Encoding Optimization (GBO) which optimizes a different number of pulses at each layer, based on our in-depth analysis that each layer has a different level of noise sensitivity. 
The proposed heterogeneous layer-wise bit encoding scheme achieves high noise robustness with low computational cost.
Our experimental results on public benchmark datasets show that GBO improves the classification accuracy by $\sim 5-40\%$ in severe noise scenarios.
\end{abstract}


\section{Introduction}

Memristive crossbars have emerged as a key component for implementing energy-efficient neural network accelerators  by performing Matrix-Vector-Multiplication (MVM) in analog domain \cite{ambrogio2018equivalent} using emerging Non-Volatile-Memory (NVM) devices.
%   such as Phase Change Memory (PCM), Resistive RAM (ReRAM),  and Spintronic devices, to store the weights of neural networks \cite{upadhyay2019emerging}.
% These NVM devices  enable low-power operation and high on-chip storage density, thereby, making the memristive crossbar a prospective computing unit for neural network accelerator.
The efficiency of the memristive crossbar can be significantly amplified by implementing Binary-Weight Neural Networks (BWNNs) \cite{courbariaux2015binaryconnect}, which consist of binary weights and multi-bit activations.
% In the last few years, Binary-Weight Neural Networks (BWNNs) \cite{courbariaux2015binaryconnect,rastegari2016xnor}, which consists of binary weights and multi-bit activations, have been proposed. 
% These networks can reduce the number of matrix-vector-multiplication operations while providing more accurate classification performance than Binary Neural Networks (BNNs) where both weights and activations are binarized. 
Encoding binary values in NVM devices is much simpler than multi-level representation, therefore, BWNNs can be easily mapped on  a binary memristive crossbar \cite{ni2017distributed}.
The multi-bit activations of BWNNs can be implemented in an efficient way by converting multi-level activation into temporal binary pulses as demonstrated in  \cite{shafiee2016isaac}. 
This approach induces less  overhead from analog-to-digital conversion and I/O than using multi-level input activation \cite{ni2017distributed}.
Overall, BWNNs  can be implemented on a binary memristive crossbar with the temporal binary bit encoding scheme.
% , as shown in Fig. \ref{fig:Xbar_concept}.
%  technique for implementing BWNNs, as shown in Fig. \ref{fig:Xbar_concept}.
% Overall, we use a binary memristive crossbar with the  temporal bit encoding technique for implementing BWNNs, as shown in Fig. \ref{fig:Xbar_concept}.
% After column-wise MVM operation at each time-step, the computation results are accumulated to obtain the final results.

% To maximize energy and area efficiency, binary memristive crossbar \cite{ni2017distributed,fouda2019mask} has been leveraged for MAC operations, which consists of binary weights and binary activations. In addition to its efficiency, the binary memristive crossbar induces less overhead from analog-to-digital conversion and I/O \cite{ni2017distributed}. Moreover, the technology for implementing binary level on memristive devices is much more mature than multi-level representation.  Nonetheless, using binary activation and weights induces a huge performance degradation in the accuracy of deep neural networks. In order to minimize the performance drop while preserving the advantage of a binary memristive crossbar, we use multi-level activation but convert it into a temporal binary pulses \cite{soliman2020ultra,shafiee2016isaac,bojnordi2016memristive}. As shown in Fig. \ref{fig:Xbar_concept}, the binary encoded pulses are taken as an input of binary memristive crossbar. After column-wise MVM operation, the temporal information is accumulated in order to obtain the computation results. 

However, due to the analog nature, the binary memristive crossbar suffers from non-idealities (\ie, noise), such as, device variations, analog-to-digital converter (ADC) and digital-to-analog converter (DAC) error, sneak path, and parasitic resistances \cite{jain2019cxdnn,bhattacharjee2021neat}. The various types of noise are amplified through multiple layers of neural network implementation on crossbars, resulting in severe performance degradation. 
To address this problem, the recent algorithm-level approaches train the weight parameters of neural networks in correspondence with the noise data.
% or pre-defined distribution from the circuit and device.
Chakraborty \textit{et al.} \cite{chakraborty2020geniex} collect  SPICE simulation data by considering both linear and non-linear noise in crossbars. 
% Lee \textit{et al.} \cite{lee2020learning} train network weights by using a surrogate model that can alleviate the time cost for conducting SPICE simulation process. 
However, it is difficult to collect noise data for each crossbar. Therefore, previous {noise-aware training} methods can be ineffective in large-volume manufacturing where each crossbar has a large device-to-device and circuit-to-circuit variation.
% these methods require a new training process for different hardware setting, therefore bring less efficiency on large-volume industrial production which has  a large device-to-device variation and circuit-to-circuit variation.
Moreover,  we assert that there is more room to improve the robustness of the memristive crossbar.
% they still cannot fully recover the clean accuracy. 
% Therefore, we assert that there is more room to improve the robustness of a crossbar.

In this paper, we introduce a new perspective on mitigating the inherent noise of a binary memristive crossbar.
We solely focus on manipulating the number of pulses of input bit encoding for a memristive crossbar.  
The proposed method can improve noise robustness significantly without weight parameter tuning, and remains compatible with previous works. Our work is based on two insights:
% We first analyze the effect of the \textcolor{red}{input pulse length} on inherent noise in crossbar. 
\textit{
(1) The inherent noise of a memristive crossbar can be mitigated by increasing the number of input pulses (see Section II-B).
(2) Each layer has a different level of noise sensitivity (see Section II-C).}
% Here, we use different encoding pulse length for different layers since each layer has the different level of noise sensitivity (observed in Section II-B). 
Considering these observations, we present Gradient-based Bit encoding Optimization (GBO) which optimizes layer-wise pulse length for input bit encoding. 
Before our main training phase, the networks are pre-trained with cross-entropy loss. 
Then, we fix the weights of networks and only train learnable parameters that represent the importance of each bit encoding strategy for each layer.
% During the training phase, we first define a pulse length set.
% Then, we assign learnable parameters for each  pulse length  strategy, which learn the importance of the corresponding strategy.
The objective function consists of two terms with distinct purposes. One is for improving classification accuracy and the other term is for  reducing the computational cost. 
The learnable parameters are optimized with two loss terms and find a saddle point.
% , while network weight parameters are fixed. 
During inference, we select the bit encoding scheme which has the maximum learnable parameter value in each layer.

% \textcolor{red}{TBD
% Also, this coarse search space can induce a sub-optimal solution for gradient-based training with the same latency budget......}

One important design parameter of GBO is the search space interval of the pulse length set.
The large interval of pulse strategies can induce a huge latency cost for the system.
Therefore, to enable a more fine-grained search space in the pulse length set, we propose Pulse Length Approximation (PLA) which allows variable fine-grained pulse length  while preserving similar information. 
We found that the activation in deep layers converges to -1 or 1   because of consecutive batch normalization \cite{ioffe2015batch} and bounded non-linear function (\eg, Tanh).
Based on this, we manipulate the pulse length by adding or subtracting pulses towards -1 or 1 values.

% There are several advantages of our method. Firstly, our proposed method can be easily compatible with the previous noise-aware training works since our work does not train the weight of networks, resulting in further performance gain.
% Secondly, our gradient-based approach can automatically find the optimal bit encoding scheme. Thus, compared to a heuristic approach (e.g., manually selecting bit encoding for each layer), our work provides general solution to various network configurations such as different datasets. 
% Moreover, our optimization process has less overhead as we train a few number of parameters. Note, we only train learnable parameters for controlling bit encoding strategy while fixing network weight parameters.

In summary, our contributions are as follows: 
(1) To the best of our knowledge, this is the first work that analyzes the effect of bit encoding on crossbar noise. 
% Our proposed method is compatible with previous noise-aware training methods as our work does not train the weight of networks.
(2) The proposed GBO can automatically find the optimal bit encoding scheme. Thus, compared to a heuristic approach (\eg, manually selecting bit encoding for each layer), our work provides a more general solution to various network configurations. 
% (3) PLA allows a fine-grained pulse length search space for GBO, preventing a memristive crossbar from having long latency. Moreover, it enhances the advantage of gradient-based training to find more optimal solution within the same computational budget. In our experiments, we show  that a fine-grained search space setting achieves higher accuracy than a coarse search space setting.
(3) Our experimental results show that GBO can  effectively mitigate crossbar noise on CIFAR10. The proposed GBO improves the classification accuracy by  $\sim 5-40\%$ in severe noise scenarios.

\section{Preliminary}

\subsection{Crossbar Array Implementation}

In this paper, we use binary weights due to their memory efficiency and the maturity of implementation on NVM devices. For activations, we encode N-bit information into sequential binary bit pulses (see Section II-B for details). Thus, our networks have multi-bit activation on a binary memristive crossbar. We use a hyperbolic tangent function as an activation function to confine the input activation range into [-1, 1]. 
% Moreover, Batch Normalization (BN) \cite{ioffe2015batch} is applied to all hidden layers, which enhances training stability and convergence. 
% In our crossbar system, we assume that a digital arithmetic unit communicates with the crossbar in order to process side operations such as bias addition and BN.
We assume that a simplified Gaussian noise is added after the MVM operation.
% This is because our objective is to show the effect of manipulating binary input encoding on crossbar noise, rather than designing an accurate noise model.
For a crossbar that stores weight matrix $W$, we can formulate an output current feature vector as:
\begin{equation}
        o = Wx + N(0,\sigma^2),
        \label{eq:noise_concept}
\end{equation}
where, $x$ is the input voltage vector, and $N(0,\sigma^2)$ is
 Gaussian noise from a memristive crossbar.

% Note that our objective is to show the effect of manipulating binary input encoding on crossbar noise, not a modeling accurate crossbar noise.

% Therfore, we assume 

% We consider two types of noise. Device variation is and IR-drop. These are regarded as the major reason to the performance degradation in crossbar \cite{he2019noise}. In order to model both types of noise, we use Gaussian noise (\ie, $N(0,\sigma_{device}^2)$  and $N(0,\sigma_{IR-drop}^2)$),  following the previous work \cite{he2019noise,zhu2020statistical}. 

% For a $m \times m$ size crossbar, given input voltage vector $x$ and weight matrix $W$, we can compute an output current feature vector as:

% \begin{equation}
% \begin{split}
%         o & = (W+N(0,\sigma_{device}^2))x + N(0,\sigma_{IR-drop}^2) \\
%         & = Wx + N(0, m\sigma_{device}^2+ \sigma_{IR-drop}^2) = Wx + N(0,\sigma^2).
% \end{split}
%         \label{eq:noise_concept}
% \end{equation}

% Here, two noise terms can be merged since $x$ consists of [-1, 1] values and we assume independent Gaussian noise. Overall, by combining the two types of noise, we can define an additive Gaussian noise $N(0,\sigma^2)$  for each crossbar.
% Note that, we use 

\begin{figure}[t!]
  \begin{center}
    \includegraphics[width=0.5\textwidth]{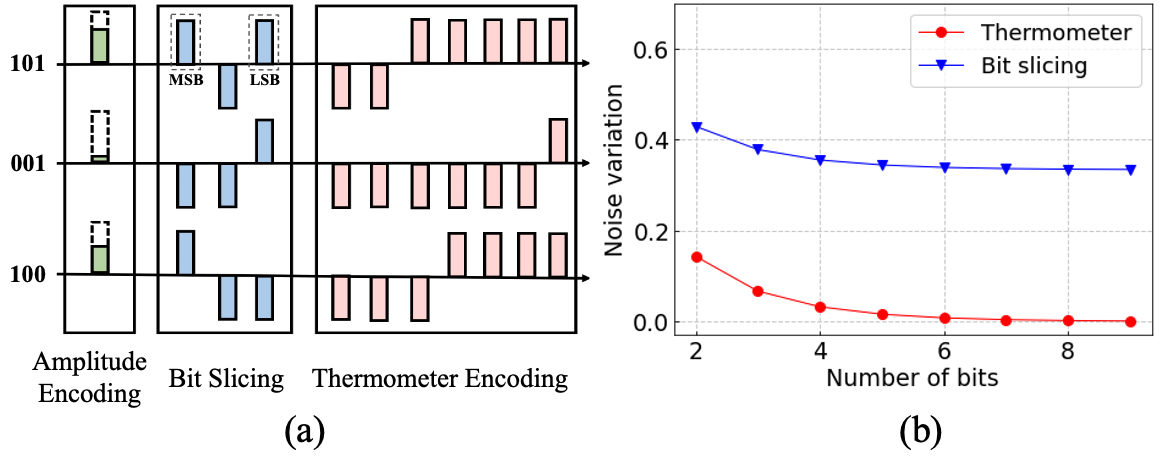}
  \end{center}
    \vspace{-5mm}
  \caption{(a) Illustration of three input bit encoding schemes.  (b) Noise variation  with respect to the number of bits. 
  Here, we set baseline noise variation as $1$.
%   \textcolor{red}{remove FIg.2(a)(b)(c) fix / reference graph(b)}
  }
   \label{fig:bitencodingschems}
     \vspace{-5mm}
\end{figure}

\subsection{Input Bit Encoding Schemes for Memristive Crossbar}

In general, the input voltage of the crossbar ($N$-bit) can be converted into pulses in different ways as shown in  Fig. \ref{fig:bitencodingschems}(a). One of the most commonly used encoding techniques is encoding an input voltage to the amplitude of $2^N$ voltage levels. However, this will require a high-precision DAC to interface with the crossbar that incurs huge area and energy consumption. Moreover, because of the wide voltage range for multi-bit representation, synapse cells should guarantee a linear I–V curve over the range of input voltage \cite{hu2016dot,chi2016prime}. 
Compared to multi-level input representation, binary input encoding brings less circuit overhead  from  analog-to-digital conversion \cite{ni2017distributed}.
\textit{Bit slicing} \cite{bojnordi2016memristive} is one of the most popular binary encoding techniques, which represents pulses with the same sequence of input bits. Unfortunately, weighted summation (based on bit position) 
induces more noise to the output current (Eq. \ref{eq:bitslicing}). To address this, we utilize \textit{Thermometer coding}  \cite{soliman2020ultra} where the number of positive pulses is proportional to a representation level.
We show that \textit{Thermometer coding} can achieve higher noise robustness than the bit slicing approach in the following sub-sections.

\textbf{Noise analysis of binary bit encoding:} We analyze the effect of two binary bit encoding schemes, \ie,  \textit{Bit slicing} and \textit{Thermometer Coding}. As we convey $b$-bits through multiple time-steps (\ie, pulses), Gaussian noise is accumulated. 
We assume that Gaussian noise is independent at each time-step. 
According to the bit encoding scheme, we reformulate the equation for MVM operation (Eq. \ref{eq:noise_concept}) with  crossbar noise.

 \textit{Bit slicing} generates $p$ pulses which have the same sequence with bit representation.
 Therefore, each pulse has a different contribution to the output result according to its bit position: 
\begin{equation}
        o = \frac{1}{\sum_{i=0}^{p-1}2^i} \sum_{i=0}^{p-1} 2^iWx_i+ N(0,\frac{\sum_{i=0}^{p-1} (2^i)^2}{(\sum_{i=0}^{p-1} (2^i))^2}\sigma^2).
    \label{eq:bitslicing}
\end{equation}
The first term represents the MVM results, and the second term shows the accumulated noise.

\textit{Thermometer coding} represents $p$-level information with $p-1$ pulses, where $p$ can be any positive integer. The output current is accumulated across $p$ pulses:
\begin{equation}
    \begin{split}
        o
        %  = \frac{1}{2p} \sum_{i=0}^{2p-1} Wx_i+ N(0,\frac{1}{2p} \sigma^2).
         = \frac{1}{p} \sum_{i=0}^{p-1} Wx_i+ N(0,\frac{1}{p}\sigma^2).
    \end{split}
    \label{eq:thermometer}
\end{equation}
% \vspace{-1mm}

\begin{figure}[t!]
  \begin{center}
    \includegraphics[width=0.31\textwidth]{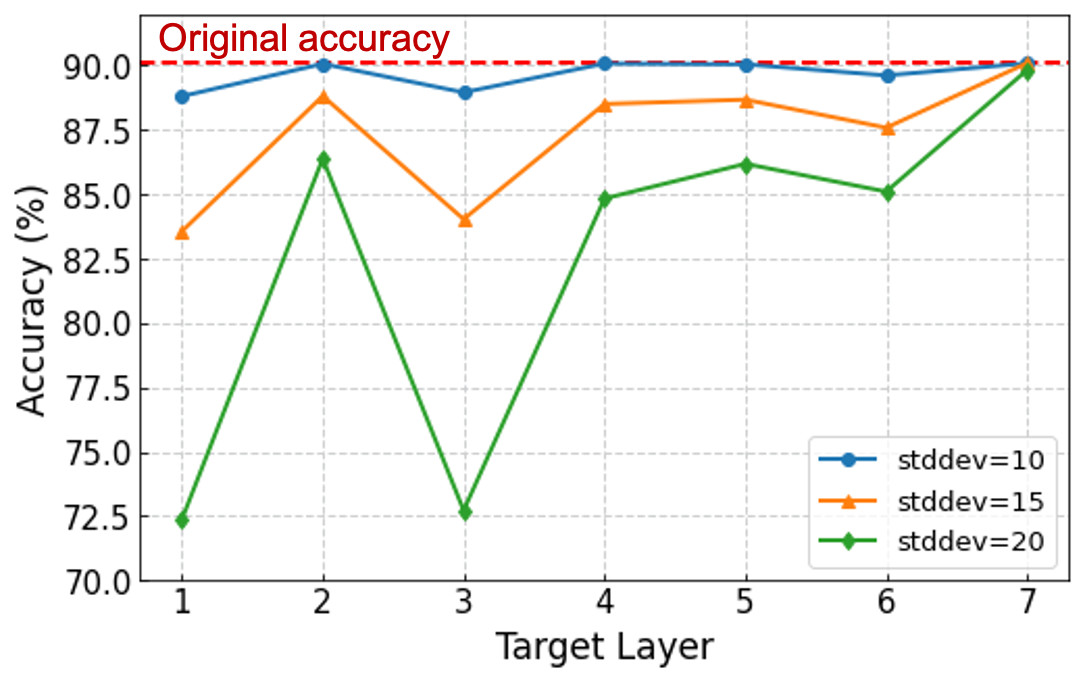}
  \end{center}
  \vspace{-5mm}
  \caption{Layer-wise noise sensitivity analysis. We use VGG9 architecture trained on CIFAR10. Target layer (x-axis) denotes the only layer where Gaussian noise $N(0,\sigma^2)$ is added.
  }
   \label{fig:layernoisesensitivity}
     \vspace{-5mm}
\end{figure}

Through mathematical analysis, we can observe the following: (1) We quantitatively compare the normalized noise variance of the two bit encoding schemes with respect to $b$-bit information. 
In Fig. \ref{fig:bitencodingschems}(b), we present the trend of noise variation with respect to the number of bits.
Here, \textit{Thermometer coding} shows higher robustness compared to \textit{Bit slicing} when they represent same bit information. 
Therefore, we use \textit{Thermometer coding} as a baseline bit encoding  scheme to build more noise robust crossbars.
% but our layer-wise coding optimization scheme can be applied to other coding schemes. 
(2) For both encoding schemes, the noise variance is inversely proportional to the number of pulses. This indicates that noise can be mitigated by increasing the number of pulses when representing the same amount of bit information. This key observation is leveraged in our layer-wise variable bit encoding optimization.

\subsection{Layer-wise Noise Sensitivity Analysis}

Uniformly increasing the latency across all layers brings longer processing time for crossbars. To efficiently take advantage of the noise suppression effect, we vary the length of bit coding in each layer according to the noise robustness. This is based on the observation that each layer has different noise sensitivity. In Fig. \ref{fig:layernoisesensitivity}, we present  noise sensitivity of each layer by adding Gaussian noise $N(0,\sigma^2)$  to the output feature map at that layer only. Noise in different layers has different impact on classification accuracy. Since different architectures have different layer configurations, a heuristic approach for selecting bit encoding for each layer is not a desirable solution. Therefore, a general framework that can support variable bit encoding scheme is required.

\vspace{-1mm}

\section{Methodology}
\vspace{-1mm}

In this section, we first introduce Gradient-based Bit encoding Optimization (GBO), which finds optimal bit encoding for a robust memristive crossbar implementation.
Then, we present Pulse Length Approximation (PLA) that enables fine-grained search space for gradient-based optimization. 

% increases or decreases the number of pulses to provide more fine-grained search space.

\vspace{-1mm}

\subsection{Learning to Optimize Input Bit Encoding Strategy}
\vspace{-1mm}

Our objective is to optimize a layer-wise bit encoding scheme by using gradient-based optimization. Let each layer $l$ of a network has a set $\Omega$ which consists of $m$ different  pulse scaling factors, thus $\Omega ^l = \{n_0^l,n_1^l,…,n_{m-1}^l\}$, $n \in \mathbb{Z}$. With the pulse scaling factor $n$, we can reformulate Eq. \ref{eq:thermometer} as follows: 
\begin{equation}
    \begin{split}
        o_{n}  = \frac{1}{np} \sum_{i=0}^{np-1} W\hat{x_i}+ N(0,\frac{1}{np}\sigma^2),
    \end{split}
    \label{eq:PLA_ntimes}
\end{equation}
where, $\hat{x_i}$ is the modified input binary pulse according to the pulse scaling factor. By doing this, we can preserve the original output while reducing noise by $1/n$. 
A conceptually naïve way to increase the number of pulses is an ensemble which repeats the same coding by $n$ times (where, $n$ is a positive integer).

For each coding scheme (\ie,  pulse scaling factor), we assign  learnable parameters $\lambda_k^l,k\in\{0,…,m-1\}$. In order to generate the probability representation of each coding scheme, we compute $\alpha_k^l$ based on $\lambda_k^l$ as $\alpha_k^l = \frac{e^{\lambda_k^l}}{\sum_{z \in \Omega} e^{\lambda_z^l}}$. 
% \begin{equation}
    % \alpha_k^l = \frac{e^{\lambda_k^l}}{\sum_{z \in \Omega} e^{\lambda_z^l}}.
% \end{equation}
% \textcolor{red}{Note that, each encoding scheme brings different approximation error from PLA (Section III-A) and noise intensity.} 
Therefore, during training, we average the impact of each coding scheme based on $\alpha_k^l$ value. We can represent an output activation as: 

\begin{equation}
    o^l = Wo^{l-1} + \sum_{k} \alpha^l_k N(0, \frac{1}{n^l_k p} \sigma ^2).
\end{equation}
% \textcolor{red}{move epsilon to the next section PLA}
% Here, $\epsilon^l_k$ stands for the approximation error, the difference between the original signal and the approximated signal. 
For the last layer, we  calculate the classification probability using a softmax function. 
The objective function consists of two terms as follows:
\begin{equation}
    L = L_{ce}+ \gamma \sum_{l} \sum_{k} \alpha_k^l n_k^l p.
    \label{eq:overall_optimization}
\end{equation}
The first term $L_{ce}$ is the standard cross-entropy loss to maximize the classification accuracy, and the second term is a regularization term to minimize latency budget across all layers. The parameter $\gamma$ is a balancing parameter between two losses. 
Overall, by optimizing Eq. \ref{eq:overall_optimization}, the model finds a saddle point between two distinct optimization directions. The learnable parameter $\lambda_k^l$  is updated following gradient calculation:

\begin{equation}
    \Delta \lambda^l_k = \frac{\partial L}{\partial \lambda^l_k} = 
    \frac{L}{\partial o^l}
    \frac{\partial o^l}{\partial \alpha^l_k}
    \frac{\partial \alpha^l_k}{\partial \lambda^l_k}.
    \label{eq:gradient_derivation}
\end{equation}
Here, all derivative terms are continuous and differentiable. Therefore, we can update the bit encoding control parameters as  $\lambda_k^l=\lambda_k^l-\eta\Delta\lambda_k^l$, where $\eta$ denotes the learning rate. 
% This training protocol is similar to previous  gradient-based neural architecture search methods \cite{liu2018darts}.  However, our optimization objective consists of two terms with opposite objectives, whereas most previous methods optimize the networks to achieve the best accuracy with a cross-entropy loss. Thus, our work shows the feasibility of using adversarial objectives in the meta-learning framework.
During inference, we select the bit encoding strategy which has the maximum importance score $\lambda_k^l$, \ie, $    n_{optimal}^l = \argmax_{n_k \in \Omega} \lambda^l_k
$.
% \begin{equation}
    % n_{optimal}^l = \argmax_{n_k \in \Omega} \lambda^l_k.
    % \label{eq:inference}
% \end{equation}
% The overall training/inference process is summarized in Algorithm 1.
Importantly, we first train the weight parameters with cross-entropy loss. After that, we only train learnable parameters $\lambda_k^l$ from a pre-trained model. This ensures a stable convergence.
\vspace{-1mm}

\subsection{Pulse Length Approximation (PLA)}
\vspace{-1mm}

In our GBO, we use the ensemble strategy for mitigating crossbar noise.
However, the ensemble strategy is only available for positive integer $n$, which can increase latency significantly.
For example, given that we use 8-pulse \textit{Thermometer coding} (p = 8), the number of pulses of the ensemble cases can be $\{8, 16, 24, …\}$. 
% In other word, even we use only n=2, the latency is doubled.
Also, this coarse search space can induce a sub-optimal solution for gradient-based training with the same latency budget. 
Thus, there is a need to enable more fine-grained pulse search space. 
% In our experiments, we show that enabling fine-grained pulse length achieves better performance than the coarse interval setting (Section IV-D). 

To address this, we present Pulse Length Approximation (PLA). The main idea is to allow flexible pulse length (\ie, float $n$) by approximating the original pulse information. 
%
% According to the scaling factor n (in Eq. \ref{eq:PLA_ntimes}), we divide two cases, i.e., (i) Ensemble: we repeat the total pulses by n times (n is positive integer) as we discussed in the previous paragraph. This case does not induce an error. (ii) Approximation: here, n is positive float value where np is integer. For example, given that we use 8-pulses thermometer coding (p = 8), the number of pulses of the ensemble cases can be {8, 16, 24, …}. On the other hand, the approximation cases can be {1, 2, …, 7, 9, …}.  
%
Specifically, we approximate the modified input signal $\hat{x_i}$  towards -1 or 1 according to its sign. This strategy is easy to be implemented by adding or removing pulses from the original pulse code. 
Our PLA is based on the observation that the activation in deep layers converges to [-1, 1].
This is because a BN layer span the range of activation by normalization, and a bounded non-linear activation function (\eg, Tanh)  limits the range of activation into [-1, 1]. 
Note that, the approximated pulses cannot exactly represent the original representation levels that leads to approximation error. 
To empirically resolve such concern, in Table \ref{table:cifar10cifar100results}, we report the classification accuracy with respect to $n$ pulses (i.e., denoted as PLA$_{n}$). The results show that the performance degradation caused by the approximation error is negligible.

\vspace{-1mm}

\section{Experiments}
\vspace{-1mm}
\subsection{Experimental Setup}
\vspace{-1mm}
We evaluate our method on CIFAR10 with VGG 9 architecture \cite{simonyan2014very}. Our implementation is based on PyTorch. For the pre-training 
stage, we use standard SGD with momentum 0.9, weight decay 5e-4. The base learning rate is set to 1e-3 and we use step-wise learning rate scheduling with a decay factor 10 at 50$\%$, 70$\%$, and 90$\%$ of the total number of epochs. Here, we set the total number of epochs to 60 for both datasets. We also quantize the activation and weight to 9-levels and binary representation, respectively, during pre-training. Thus, activations  are  converted to the \textit{Thermometer coding} with 8 pulses. For GBO training, we use ADAM optimizer with learning rate 1e-4. We only train $\lambda_k^l$ parameter during 10 epochs. We set  pulse scaling set $\Omega$ as [0.5, 0.75, 1, 1.25, 1.5, 1.75, 2] that results in a pulse length set as  [4, 6, 8, 10, 12, 14, 16].

\begin{table}[t]
    \addtolength{\tabcolsep}{1.5pt}
    \centering
    \caption{Results on CIFAR10   with VGG9 architecture}
    \label{table:cifar10cifar100results}
    \resizebox{0.50\textwidth}{!}{%
    \begin{tabular}{lcccc}
        \toprule
        Method 
        & Noise $\sigma$  &
          \# pulses  in each layer  &
         Avg.\# pulses   
         & Acc. (\%)  \\
        \midrule
        Baseline  \hspace{10mm}  &   10 & [8, 8, 8, 8, 8, 8, 8] & 8 & 83.94 \\
        PLA$_{10}$ &    10 & [10, 10, 10, 10, 10, 10, 10] & 10 & 85.38 \\
        PLA$_{12}$ &     10 & [12, 12, 12, 12, 12, 12, 12] & 12 & 85.58 \\
        PLA$_{14}$ &   10 & [14, 14, 14, 14, 14, 14, 14] & 14 & 86.24 \\
        PLA$_{16}$ &     10 & [16, 16, 16, 16, 16, 16, 16] & 16 & 88.27 \\
        GBO ($\sim$PLA$_{10}$) &     10 & [10, 10, 8, 10, 10, 4, 6] & 9.71 & 86.36 \\
        GBO ($\sim$PLA$_{14}$) &      10 & [16, 6, 12, 6, 10, 14, 16] & 14.85 & 88.27
        \\
        \midrule
        Baseline &     15 & [8, 8, 8, 8, 8, 8, 8] & 8 & 62.27 \\
        PLA$_{10}$ &      15 & [10, 10, 10, 10, 10, 10, 10] & 10 & 71.09 \\
        PLA$_{12}$ &    15 & [12, 12, 12, 12, 12, 12, 12] & 12 & 74.61 \\
        PLA$_{14}$ &     15 & [14, 14, 14, 14, 14, 14, 14] & 14 & 77.53 \\
        PLA$_{16}$ &      15 & [16, 16, 16, 16, 16, 16, 16] & 16 & 82.95 \\
        GBO ($\sim$PLA$_{10}$) &    15 & [8, 12, 8, 10, 14, 14, 8] & 10.42 & 76.35 \\
        GBO ($\sim$PLA$_{14}$) &      15 & [16, 16, 14, 16, 16, 16, 6] & 14.28 & 82.73 \\
          \midrule
        Baseline  &    20 & [8, 8, 8, 8, 8, 8, 8] & 8 & 31.46 \\
        PLA$_{10}$ &    20 & [10, 10, 10, 10, 10, 10, 10] & 10 & 42.94 \\
        PLA$_{12}$ &    20 & [12, 12, 12, 12, 12, 12, 12] & 12 & 51.89 \\
        PLA$_{14}$ &    20 & [14, 14, 14, 14, 14, 14, 14] & 14 & 58.80 \\
        PLA$_{16}$ &     20 & [16, 16, 16, 16, 16, 16, 16] & 16 & 67.49 \\
        GBO ($\sim$PLA$_{10}$) &     20 & [12, 10, 8, 8, 14, 14, 6] &  10.28 & 46.33 \\
        GBO ($\sim$PLA$_{14}$) &      20 & [16, 16, 16, 14, 14, 14, 12] & 14.57 & 71.53 \\
          \bottomrule
    \end{tabular}%
    \vspace{-8mm}
    }
\end{table}

\vspace{-1mm}

\subsection{Experimental Results}
\vspace{-1mm}

% \textcolor{red}{require more explanatrion for notation and Fig5 Fig6}
In Table \ref{table:cifar10cifar100results}, we present the performance of baseline, PLA, and GBO. $PLA_n$ denotes PLA with $n$ pulses.
We set $\sigma$ in noise modeling as [10, 15, 20], which can cover the wide range of noise levels in a crossbar. 
We obtain 90.80\% accuracy without considering crossbar noise.
Note that we can obtain different bit encoding solution based on trade-off parameter $\gamma$ in Eq. \ref{eq:overall_optimization}.
In the table, we report two GBO cases that have  similar latency with $PLA_{10}$ and $PLA_{14}$. 
From Table \ref{table:cifar10cifar100results}, we can observe the following: (1) PLA improves the performance for all noise scenarios. This shows that crossbar noise can be mitigated by increasing the number of pulses. (2) GBO enables heterogeneous bit encoding strategies. According to the optimization results, we present the number of pulses in each layer and averaged pulse number. The results show that GBO achieves better accuracy compared to PLA with similar number of pulses. 

\vspace{-1mm}

\subsection{Synergy with the Previous Noise-aware Training }
\vspace{-1mm}

% The most previous methods for a noise-less crossbar train the network weights using noise data or noise distribution \cite{chakraborty2020geniex, lee2020learning, he2019noise}. 

In this experiment, we show the synergy effect between GBO and Noise-Injection Adaptation (NIA) \cite{he2019noise}. As shown in Table \ref{table:synergy_effect}, compared to \textit{Baseline}, using both GBO and NIA together outperform the baseline in terms of accuracy. Note that, GBO aims to only manipulate the input encoding scheme, therefore it is natural that the overall performance gain is lower than fine-tuning weight parameters which can represent sophisticated noise distribution. Combining two techniques brings  further performance gain for all possible noise scenarios. 
The results show that GBO reduces the impact of noise intensity (\ie, noise deviation), enhancing the effect of noise-aware training. 
% Moreover, applying PLA also shows a similar effect with GBO. Overall, GBO can be combined with the previous noise-aware training approaches.

% \vspace{-1mm}
% \subsection{Ablation Studies on Search Grid Interval}
% \vspace{-1mm}

% In our method, we define the set $\Omega$ which consists of $m$ different scaling factors for a variable pulse length. According to the elements of set $\Omega$, the optimization results are likely to be varied. In order to investigate the effect of search grid interval of $\Omega$, in Fig. \ref{fig:noise_intensity}(b), we compare the performance on \textit{interval-1} (dense search space), \textit{interval-2} (our original setting), and \textit{interval-4} (coarse search space). 
% Thus, the pulse sets for three configurations are [4, 5, 6, ..., 15, 16], [4, 6, 8, ..., 14, 16], [4, 8, 12, 16], respectively.
% For all scenarios, we sweep the $\gamma$ parameter in order to cover pulse length between 9 and 16. The results show \textit{interval-1} achieves the best performance for all pulse lengths. This implies that a small interval setting enables a fine-grained bit encoding configuration, therefore resulting in higher accuracy than the coarse interval settings.
% \vspace{-1mm}

\vspace{-2mm}
\section{Conclusion}
\vspace{-1mm}
For the first time, we explore the impact of binary encoding techniques on crossbar noise.
Interestingly, we found that the noise can be mitigated by increasing the length of bit encoding.
Based on this observation, we propose Gradient-based Bit encoding Optimization (GBO) which enables heterogeneous bit encoding schemes across all layers.
Through experiments, we observe that GBO effectively addresses the  inherent noise problem.
Moreover, combining GBO with the previous noise-aware training (\ie,  NIA) shows a synergy effect in terms of accuracy improvement.

\begin{table}[t]
    \addtolength{\tabcolsep}{1.5pt}
    \centering
    \caption{Synergy effect with the noise-aware training.}
    \vspace{-1mm}
    \label{table:synergy_effect}
    \resizebox{0.41\textwidth}{!}{%
    \begin{tabular}{lccc}
        \toprule
        Method \\ (Acc / avg. \#pulses) &  $\sigma = 10$  &  $\sigma = 15$ &  $\sigma = 20$ \\
        \midrule
        Baseline & 83.94 / 8 & 62.27 / 8 & 31.46 / 8 \\
        NIA \cite{he2019noise} & 88.35 / 8 & 84.84 / 8 & 78.78 / 8 \\
        GBO & 86.36 / 9.71  & 76.35 / 10.21 & 46.33 / 10.28 \\
        NIA \cite{he2019noise} + GBO & 88.93 / 9.71 & 86.45 / 10.24 & 81.33 / 10.28 \\
        NIA \cite{he2019noise} + PLA  & 88.91 / 10 & 85.17 / 10 & 80.29 / 10 \\
        \bottomrule
    \end{tabular}%
    % \vspace*{0.15in}
    }
        \vspace{-4mm}
\end{table}

\vspace{-2mm}

\section*{Acknowledgement}
\vspace{-1mm}
This work was carried out while Youngeun Kim worked as an intern at Samsung Advanced Institute of Technology, South Korea.
The research was funded in part by C-BRIC, one of six centers in JUMP, a Semiconductor Research Corporation (SRC) program sponsored by DARPA, and the National Science Foundation (Grant\#1947826).

% \begin{figure}[t!]
%   \begin{center}
%     \includegraphics[width=0.50\textwidth]{figures/sweep_all_4p.png}
%   \end{center}
%   \vspace{-5mm}
%   \caption{Comparing accuracy and the number of pulses  across baseline, GBO, and PLA. 
%   For GBO, we change  the $\gamma$ hyperparameter to cover a similar pulses configuration with PLA.
%   }
%     \vspace{-3mm}
%   \label{fig:sweep_time_acc}
% \end{figure}

% \begin{figure}[t]
% \begin{center}
% \def\arraystretch{0.5}
% \begin{tabular}{@{}c@{\hskip 0.01\linewidth}c@{}c}
% \includegraphics[width=0.49\linewidth]{figures/robust_cifar10.png} &
% \includegraphics[width=0.46\linewidth]{figures/searchgrid_ablation.png} 
% \\
% % \vspace*{0.1in}
% {\hspace{1mm} (a) } & {\hspace{1mm} (b) }\\
% \end{tabular}
% \end{center}
%   \vspace{-3mm}
% \caption{ (a) The performance change of baseline, PLA, and GBO with respect to the noise intensity. 
% (b) Performance and latency change with respect to search grid interval. We use a CIFAR10 dataset.
% }
%   \vspace{-3mm}
% \label{fig:noise_intensity}
% \end{figure}

% \begin{figure}[t!]
%   \begin{center}
%     \includegraphics[width=0.25\textwidth]{figures/searchgrid_ablation.png}
%   \end{center}
%   \vspace{-5mm}
%   \caption{Performance and latency change with respect to search grid interval. We use a CIFAR10 dataset.
%   }
%   \label{fig:searchgrid_ablation}
%      \vspace{-3mm}
% \end{figure}

% \vspace{-1mm}
{\tiny
\vspace{-1mm}
\bibliographystyle{ieee_fullname}
\bibliography{egbib}

\begin{thebibliography}{10}\itemsep=-1pt

\bibitem{ambrogio2018equivalent}
Stefano Ambrogio et~al.
\newblock Equivalent-accuracy accelerated neural-network training using
  analogue memory.
\newblock {\em Nature}, 558(7708):60--67, 2018.

\bibitem{bhattacharjee2021neat}
Abhiroop Bhattacharjee et~al.
\newblock Neat: Non-linearity aware training for accurate, energy-efficient and
  robust implementation of neural networks on 1t-1r crossbars.
\newblock {\em IEEE Transactions on Computer-Aided Design of Integrated
  Circuits and Systems}, 2021.

\bibitem{bojnordi2016memristive}
Mahdi~Nazm Bojnordi and Engin Ipek.
\newblock Memristive boltzmann machine: A hardware accelerator for
  combinatorial optimization and deep learning.
\newblock In {\em 2016 IEEE International Symposium on High Performance
  Computer Architecture (HPCA)}, pages 1--13. IEEE, 2016.

\bibitem{chakraborty2020geniex}
Indranil Chakraborty et~al.
\newblock Geniex: A generalized approach to emulating non-ideality in
  memristive xbars using neural networks.
\newblock In {\em 2020 57th ACM/IEEE Design Automation Conference (DAC)}, pages
  1--6. IEEE, 2020.

\bibitem{chi2016prime}
Ping Chi et~al.
\newblock Prime: A novel processing-in-memory architecture for neural network
  computation in reram-based main memory.
\newblock {\em ACM SIGARCH Computer Architecture News}, 44(3):27--39, 2016.

\bibitem{courbariaux2015binaryconnect}
Matthieu Courbariaux, Yoshua Bengio, and Jean-Pierre David.
\newblock Binaryconnect: Training deep neural networks with binary weights
  during propagations.
\newblock In {\em Advances in neural information processing systems}, pages
  3123--3131, 2015.

\bibitem{he2019noise}
Zhezhi He et~al.
\newblock Noise injection adaption: End-to-end reram crossbar non-ideal effect
  adaption for neural network mapping.
\newblock In {\em Proceedings of the 56th Annual Design Automation Conference
  2019}, pages 1--6, 2019.

\bibitem{hu2016dot}
Miao Hu et~al.
\newblock Dot-product engine for neuromorphic computing: Programming 1t1m
  crossbar to accelerate matrix-vector multiplication.
\newblock In {\em 2016 53nd ACM/EDAC/IEEE Design Automation Conference (DAC)},
  pages 1--6. IEEE, 2016.

\bibitem{ioffe2015batch}
Sergey Ioffe and Christian Szegedy.
\newblock Batch normalization: Accelerating deep network training by reducing
  internal covariate shift.
\newblock In {\em International conference on machine learning}, pages
  448--456. PMLR, 2015.

\bibitem{jain2019cxdnn}
Shubham Jain and Anand Raghunathan.
\newblock Cxdnn: Hardware-software compensation methods for deep neural
  networks on resistive crossbar systems.
\newblock {\em ACM Transactions on Embedded Computing Systems (TECS)},
  18(6):1--23, 2019.

\bibitem{ni2017distributed}
Leibin Ni et~al.
\newblock Distributed in-memory computing on binary rram crossbar.
\newblock {\em ACM Journal on Emerging Technologies in Computing Systems
  (JETC)}, 13(3):1--18, 2017.

\bibitem{shafiee2016isaac}
Ali Shafiee et~al.
\newblock Isaac: A convolutional neural network accelerator with in-situ analog
  arithmetic in crossbars.
\newblock {\em ACM SIGARCH Computer Architecture News}, 44(3):14--26, 2016.

\bibitem{simonyan2014very}
Karen Simonyan and Andrew Zisserman.
\newblock Very deep convolutional networks for large-scale image recognition.
\newblock {\em arXiv preprint arXiv:1409.1556}, 2014.

\bibitem{soliman2020ultra}
T Soliman et~al.
\newblock Ultra-low power flexible precision fefet based analog in-memory
  computing.
\newblock In {\em 2020 IEEE International Electron Devices Meeting (IEDM)},
  pages 29--2. IEEE, 2020.

\end{thebibliography}
}

\end{document}